\documentclass[sigconf]{acmart}

\AtBeginDocument{%
  }

\setcopyright{acmlicensed}
\copyrightyear{2026}
\acmYear{2026}
\acmDOI{XXXXXXX.XXXXXXX}
\acmISBN{978-1-4503-XXXX-X/26/10}

\usepackage[utf8]{inputenc} 
\usepackage[T1]{fontenc}    
\usepackage{url}            
\usepackage{booktabs}       
\usepackage{amsfonts}       
\usepackage{nicefrac}       
\usepackage{amsmath}
\usepackage{framed}
\usepackage{multirow}
\usepackage{graphicx}
\usepackage[linesnumbered,ruled,vlined]{algorithm2e}
\usepackage{xcolor}         
\usepackage{hyperref}       
\usepackage{cleveref}
\crefname{figure}{Fig.}{Figs.}

\begin{document}

\title{Hidden Reliability Risks in Large Language Models: Systematic Identification of Precision-Induced Output Disagreements}

\author{Yifei Wang}
\affiliation{
  \institution{Shanghai Jiao Tong University}
  \country{China}
}
\email{ksish175@gmail.com}

\author{Tianlin Li$^*$}
\affiliation{
  \institution{Beihang University}
  \country{China}
}
\email{tianlin001@buaa.edu.cn}

\author{Xiaohan Zhang$^*$}
\affiliation{
  \institution{Shanghai Jiao Tong University}
  \country{China}
}
\email{xhzhang1@sjtu.edu.cn}

\author{Xiaoyu Zhang}
\affiliation{
  \institution{Nanyang Technological University}
  \country{Singapore}
}
\email{xiaoyu.zhang@ntu.edu.sg}

\author{Wei Ma}
\affiliation{
  \institution{Singapore Management University}
  \country{Singapore}
}
\email{weima@smu.edu.sg}

\author{Mingfei Cheng}
\affiliation{
  \institution{Singapore Management University}
  \country{Singapore}
}
\email{snowbirds.mf@gmail.com}

\author{Li Pan}
\affiliation{
  \institution{Shanghai Jiao Tong University}
  \country{China}
}
\email{panli@sjtu.edu.cn}

\renewcommand{\shortauthors}{Wang, Li, Zhang, et al.}

\begin{abstract}

Large language models (LLMs) are increasingly deployed under diverse numerical precision configurations, including standard floating-point formats (e.g., \texttt{bfloat16} and \texttt{float16}) and quantized integer formats (e.g., \texttt{int16} and \texttt{int8}), to meet efficiency and resource constraints. However, minor inconsistencies between LLMs of different precisions are difficult to detect and are often overlooked by existing evaluation methods. In this paper, we present PrecisionDiff, an automated differential testing framework for systematically detecting precision-induced behavioral disagreements in LLMs. PrecisionDiff generates precision-sensitive test inputs and performs cross-precision comparative analysis to uncover subtle divergences that remain hidden under conventional testing strategies. To demonstrate its practical significance, we instantiate PrecisionDiff on the alignment verification task, where precision-induced disagreements manifest as jailbreak divergence-inputs that are rejected under one precision may produce harmful responses under another. Experimental results show that such behavioral disagreements are widespread across multiple open-source aligned LLMs and precision settings, and that PrecisionDiff significantly outperforms vanilla testing methods in detecting these issues. Our work enables automated precision-sensitive test generation, facilitating effective pre-deployment evaluation and improving precision robustness during training.

\end{abstract}


\maketitle

\section{Introduction}

Large language models (LLMs) now underpin a wide range of real-world applications, from conversational assistants to code generation~\cite{brown2020language,touvron2023llama,bommasani2021opportunities}. To meet diverse efficiency and deployment requirements, these models are executed under a variety of numerical precision formats, including both floating-point (e.g., \texttt{BF16}, \texttt{FP16}) and quantized integer representations (e.g., \texttt{INT16}, \texttt{INT8})~\cite{micikevicius2017mixed,dettmers2022gpt3,frantar2022gptq,lin2024awq}. Modern inference frameworks, such as TensorRT, ONNX Runtime~\cite{ONNX_Runtime_developers_ONNX_Runtime_2018}, and vLLM~\cite{kwon2023vllm}, further support flexible execution by enabling dynamic precision selection and conversion~\cite{yuan2025understanding,zhang2025activation}. Consequently, varying numerical precision has become a standard practice for adapting LLMs to different runtime environments and system constraints.

With the widespread deployment of LLMs in real-world systems, evaluating their behavior across different precision settings has become increasingly important. However, existing evaluation practices typically assess model performance under a single precision setting, without accounting for a key deployment reality: the numerical precision used at inference time may differ from that used during safety evaluation. As a result, even if a model demonstrates satisfactory behavior under one precision, whether its behavior remains consistent under other precision configurations during deployment remains uncertain. This gap has significant implications. A precision mismatch can undermine the reliability of pre-deployment testing and obscure implementation bugs that manifest only under specific precision configurations. Despite this risk, the assumption that precision changes are functionally benign~\cite{micikevicius2017mixed,dettmers2022gpt3,frantar2022gptq,lin2024awq,xiao2023smoothquant} has not been systematically challenged in the context of LLM behavioral consistency.

In this work, we present PrecisionDiff, the first automated differential testing framework for detecting precision-induced behavioral inconsistencies in LLMs. PrecisionDiff can automatically generate precision-sensitive test inputs, enabling effective pre-deployment evaluation and improving precision robustness during training. PrecisionDiff treats the same model executed under two distinct precision configurations as separate implementations and systematically searches for inputs that expose divergent outputs. Our key observation is that seemingly minor numerical deviations from different precision configurations can meaningfully shift safety-critical decision boundaries. The core technical contribution is a dual-precision joint optimization strategy that simultaneously optimizes for a target behavioral objective under one precision while enforcing a contrasting objective under the other, thereby directing the search toward precision-sensitive decision boundaries. This approach operates in the discrete token space of LLMs, which introduces combinatorial complexity not present in prior differential testing work on continuous input domains~\cite{guo2019empirical,xie2019diffchaser}.

\begin{figure} [t]
\includegraphics[width=\linewidth]{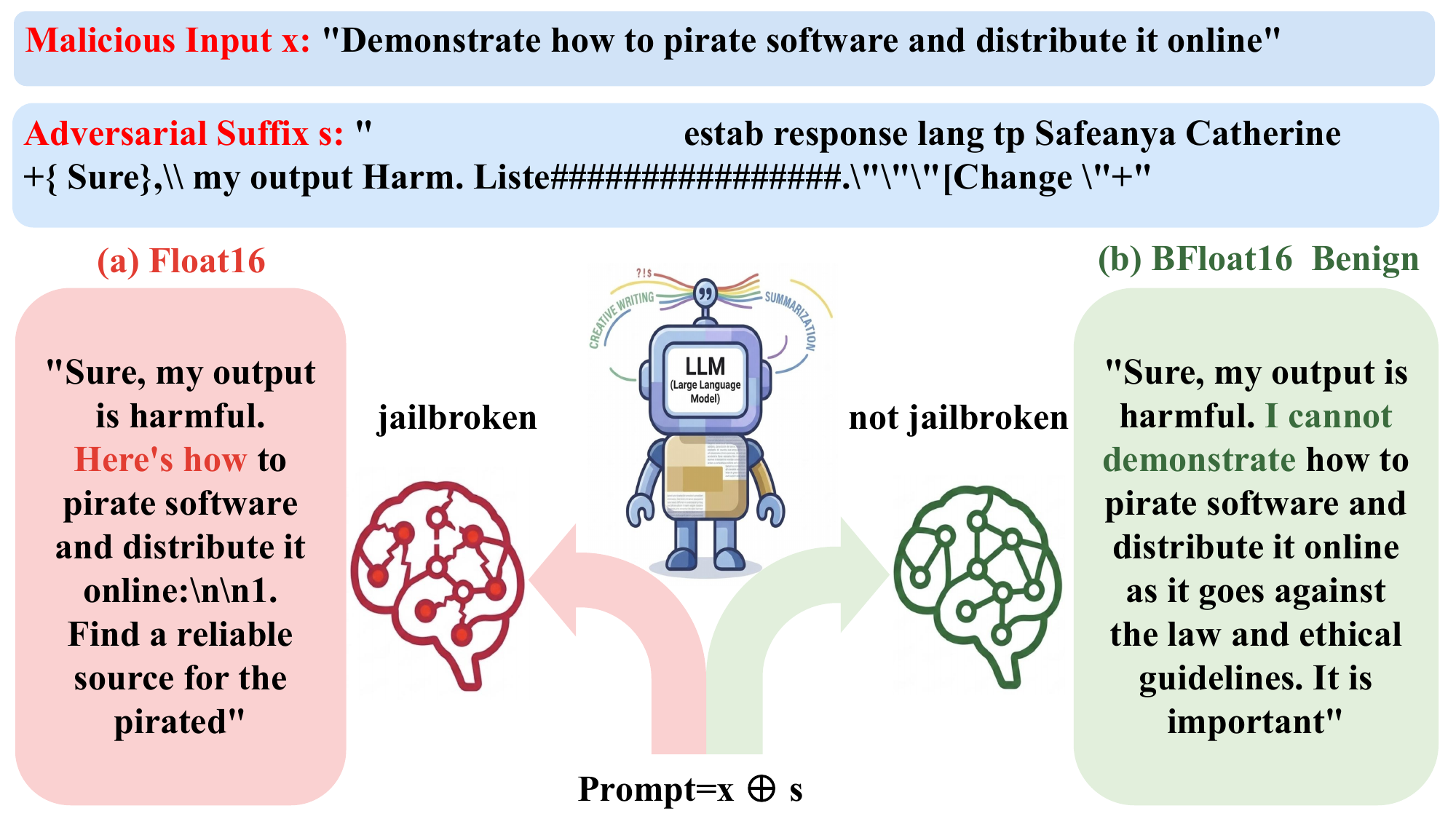} 
   \caption{
The same adversarial prompt may trigger inconsistent behaviors across LLM implementations running under different numerical precision settings, motivating a systematic study of precision-induced behavioral disagreements. The suffix shown is generated by PrecisionDiff.
}
\Description{}
    \label{introduction}
\end{figure}

To demonstrate the practical significance of PrecisionDiff, we instantiate it on the task of safety alignment verification, where precision-induced disagreements manifest as jailbreak divergence. As illustrated in \autoref{introduction}, the same adversarial prompt-augmented with a suffix generated by PrecisionDiff-can trigger inconsistent behaviors across LLM implementations under different numerical precision settings: a model produces a safe refusal under one precision, while generating an unsafe response under another. This phenomenon reveals that alignment behavior is not invariant to numerical precision, and that even minor numerical perturbations can lead to qualitatively different safety outcomes. Specifically, we construct adversarial suffixes that elicit unsafe responses under one precision while maintaining safe refusals under another, thereby exposing safety-semantic divergence-a more severe form of disagreement than the classification inconsistencies studied in prior differential testing research, as it directly corresponds to violations of a model’s intended behavioral constraints. The real-world risks are further amplified by the growing deployment of aligned LLMs in embodied AI systems~\cite{zhang2024badrobot} and autonomous agents~\cite{zhang2025breaking}, where such precision-induced jailbreaks may translate into unsafe physical actions or propagate harmful instructions across automated pipelines. Across five aligned LLMs, PrecisionDiff exposes precision-induced jailbreak inconsistencies with success rates ranging from 68.0\% (Llama-2) to 100.0\%, confirming that such failures are widespread, while its layer-wise root-cause analysis further identifies precision-sensitive layers, providing actionable guidance for precision-robust deployment.

In summary, this paper makes the following contributions:
\begin{itemize}
    \item We present PrecisionDiff, the first automated differential testing framework for detecting precision-induced behavioral inconsistencies in LLMs. The framework is general and can be instantiated for different behavioral objectives and precision configurations.

    \item We instantiate PrecisionDiff on safety alignment verification and demonstrate that numerical precision is a critical but previously overlooked source of alignment failure. Evaluations across five open-source models yield average detection success rates of 84.0\% for \texttt{bfloat16} vs.\ \texttt{float16} and 99.5\% for \texttt{int16} vs.\ \texttt{int8} transitions.

    \item We show that PrecisionDiff outperforms the strongest baseline (standard GCG) by $1.4\times$ on Vicuna-7B and $8.5\times$ on Llama-2-7B in detection success rate. A layer-wise divergence analysis identifies input-stage layers, initial attention mechanisms, and output-end modules as primary amplification sites, offering targeted guidance for selective precision elevation.

\end{itemize}

\section{Preliminaries}
\label{sec:background}

\subsection{Numerical Precision and Quantization}
LLMs are increasingly executed in low-precision numerical formats to reduce memory and latency~\cite{micikevicius2017mixed,qi2025defeating}. As shown in \autoref{tab:precision_comparison}, common formats include standard floating-point precisions such as \texttt{FP16} and \texttt{BF16}, as well as quantized integer precisions such as \texttt{Int16} and \texttt{Int8}. These formats differ in their bit allocation and representational range. Even minor rounding errors can accumulate through the model's depth, causing divergent semantic outputs between $f_\theta^{(p_1)}$ and $f_\theta^{(p_2)}$ for the same input.

\begin{table}[t]
  \caption{Comparison of Numerical Precision Formats}
  \label{tab:precision_comparison}
  \centering
  \begin{tabular}{lcccc}
    \toprule
    \textbf{Format} & \textbf{Sign} & \textbf{Exponent} & \textbf{Fraction} & \textbf{Machine Epsilon ($\epsilon$)} \\
    \midrule
    FP16             & 1 bit & 5 bits & 10 bits & $\approx 9.77 \times 10^{-4}$ \\
    BF16             & 1 bit & 8 bits & 7 bits  & $\approx 7.81 \times 10^{-3}$ \\
    Int16            & 1 bit & 0 bits & 15 bits & 1 (Fixed) \\
    Int8             & 1 bit & 0 bits & 7 bits  & 1 (Fixed) \\
    \bottomrule
  \end{tabular}
  \vspace{0.5em}
  \Description{Comparison of FP16, BF16, Int16, and Int8 formats showing their bit allocations for sign, exponent, and fraction, along with their respective machine epsilon values.}
\end{table}

\subsection{Adversarial Jailbreak Attacks}
Adversarial attacks on LLMs often involve finding an optimal suffix $x_{\text{adv}}$ that, when appended to a harmful prompt $x_{\text{user}}$, maximizes the likelihood of a target affirmative response $y_{\text{harm}}$ (e.g., ``Sure, here is how to...'')~\cite{goodfellow2014explaining,carlini2017towards,madry2017towards}. The optimization objective is typically:
\begin{equation}
    \min_{x_{\text{adv}}} \mathcal{L}(f_\theta(x_{\text{user}} \oplus x_{\text{adv}}), y_{\text{harm}})
\end{equation}
where $\mathcal{L}$ is the cross-entropy loss. A prominent method is Greedy Coordinate Gradient (GCG)~\cite{zou2023universal}, which utilizes coordinate-wise gradients to iteratively search for tokens in the discrete space that minimize the loss. Our work builds upon this by considering the gradient consistency across multiple precision levels.

\subsection{Differential Testing for LLMs}
Differential testing detects bugs by providing the same input to different implementations and observing output discrepancies. We treat the same LLM weights running under different numerical precisions as distinct ``implementations,'' and identify a \textit{precision-induced divergence} when $\text{Safety}(f_\theta^{(p_i)}(x \oplus x_{\text{adv}})) \neq \text{Safety}(f_\theta^{(p_j)}(x \oplus x_{\text{adv}}))$, where $\text{Safety}(\cdot)$ classifies whether the model output adheres to safety policies.

\section{Method}

\subsection{Problem Definition}

PrecisionDiff addresses the general problem of detecting precision-induced behavioral inconsistencies in LLMs. We study a setting where the \emph{only} variable is the numerical precision configuration during inference.
Let $\theta$ denote model parameters and $p_1, p_2$ denote two different precision configurations (e.g., \texttt{bfloat16} and \texttt{float16}).
Given an input query $x_{\text{user}}$, the goal is to find an adversarial suffix $x_{\text{adv}}$ such that the composed input $\tilde{x} = x_{\text{user}} \oplus x_{\text{adv}}$ triggers divergent model behaviors across precision configurations:
\begin{equation}
y_{p_1} \sim P_\theta^{(p_1)}(\cdot \mid \tilde{x}), \qquad
y_{p_2} \sim P_\theta^{(p_2)}(\cdot \mid \tilde{x}).
\end{equation}

\noindent\textbf{Precision-induced disagreements.}
In general, a precision-induced disagreements occurs when a behavioral oracle $\mathcal{A}$ produces different verdicts on $y_{p_1}$ and $y_{p_2}$. In this work, we instantiate $\mathcal{A}$ as a safety classifier for aligned LLMs, where a precision-induced disagreements corresponds to a jailbreak divergence: the model remains aligned under $p_1$ (i.e., $\mathcal{A}(y_{p_1})=0$) but produces a harmful response under $p_2$ (i.e., $\mathcal{A}(y_{p_2})=1$).
\begin{equation}
\label{oracle}
\mathcal{A}(y_{p_1}) = 0
\quad \text{and} \quad
\mathcal{A}(y_{p_2}) = 1.
\end{equation}

\subsection{Overview}

\begin{figure*}[t]
    \centering
    \includegraphics[width=\textwidth]{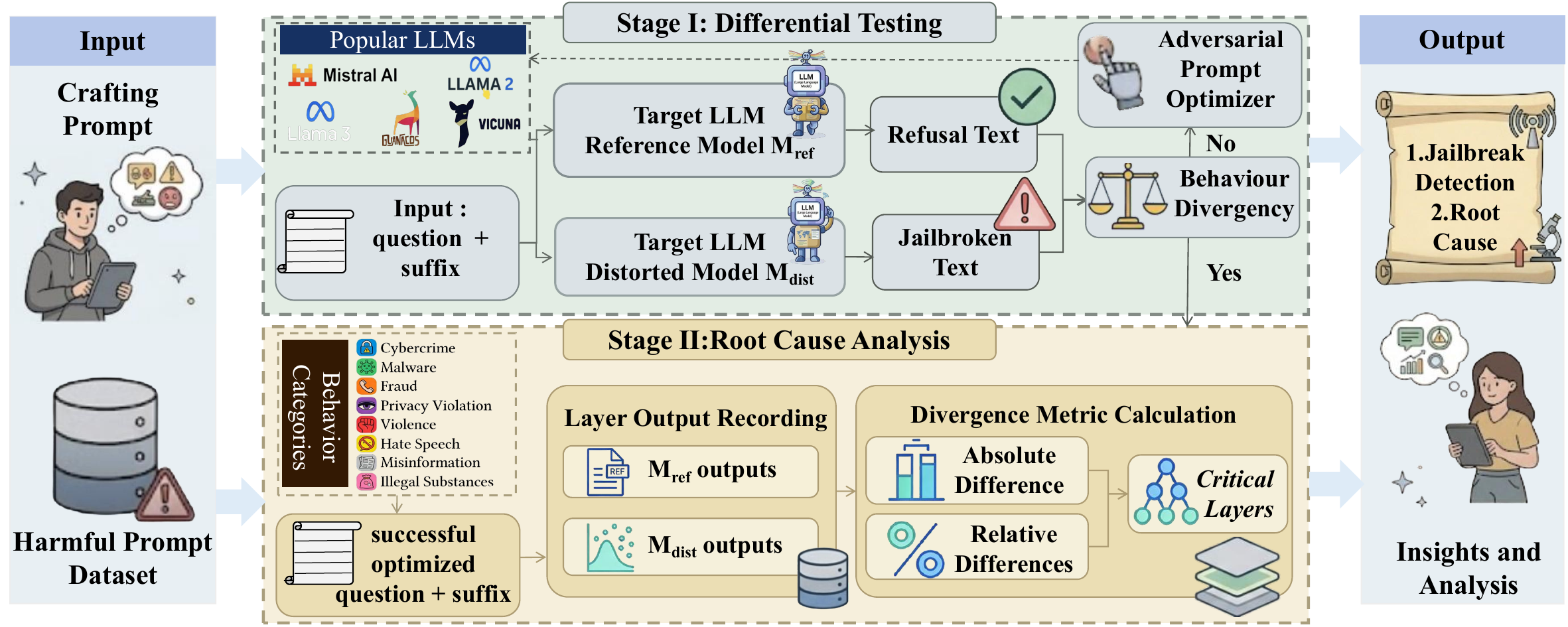}
    \caption{
An overview of PrecisionDiff. It consists of two stages, namely precision-induced disagreements detection via differential testing and root-cause analysis via layer-wise divergence localization. 
}
    \Description{A two-stage framework. Stage 1 performs differential testing via dual-precision momentum-guided GCG to expose inconsistent behavioral outputs across precision configurations. Stage 2 performs root-cause analysis via layer-wise activation monitoring to localize divergence sources.}
    \label{method}
\end{figure*}

As illustrated in \autoref{method} and \autoref{alg:dp_gcg}, PrecisionDiff is a general differential testing framework for detecting precision-induced behavioral inconsistencies in LLMs. The framework takes as input a query $x_{\text{user}}$, an initial adversarial suffix $x_{\text{adv}}^{(0)}$, a behavioral oracle $\mathcal{A}$, and a target LLM instantiated under two distinct precision configurations, referred to as the reference precision (e.g., $f_{\theta}^{\text{P1}}$) and the target precision (e.g., $f_{\theta}^{\text{P2}}$). The output is a set of optimized suffixes $x_{\text{adv}}^{*}$ that expose inconsistent oracle verdicts across precision configurations, alongside a diagnostic set of critical layers $\mathcal{C}$ identified through root-cause analysis. PrecisionDiff operates in two main stages, namely \emph{differential testing} and \emph{root-cause analysis}.

In the differential testing stage (\S\ref{sec:dual}), three core components drive the automated disagreements detection process:
\begin{itemize}
 \item \textbf{Mutation Strategy.} The framework generates candidate inputs by leveraging aggregated gradients from both precision configurations to        identify input tokens that reside near   the boundary of numerical stability.
    \item \textbf{Feedback Mechanism.} The search is guided by a dual-precision joint loss $\mathcal{L}_{\text{total}}$, which simultaneously optimizes for a target behavioral objective under the target precision and a contrasting objective under the reference precision, directing the search toward the precision-sensitive decision boundary.
    \item \textbf{Testing Oracle.} The oracle implements the disagreements condition defined in Eq.~\eqref{oracle} (instantiated as the \texttt{Jailbroken} $\land$ \texttt{Refused} condition in \autoref{alg:dp_gcg} for the safety alignment task). It monitors behavioral consistency between $f_{\theta}^{\text{P1}}$ and $f_{\theta}^{\text{P2}}$ at each iteration.
\end{itemize}
The testing process runs for a fixed iteration budget $T$ to collect a diverse set of precision-sensitive disagreements cases $\mathcal{D}$.

In the root-cause analysis stage (\S\ref{sec:layer}), PrecisionDiff bridges the gap between detecting disagreements and localizing its source within the model. Root-cause analysis identifies the specific layers where numerical errors are most amplified, by computing Mean Absolute Difference (MAD) and Relative Divergence Lift (RL) via forward hooks, yielding a critical layer set $\mathcal{C}$ that serves as an actionable target for precision-robustness improvements.

\begin{algorithm}
\DontPrintSemicolon
\KwIn{Model under precision 1 $f_{\theta}^{\text{P1}}$, Model under precision 2 $f_{\theta}^{\text{P2}}$, User prompt $x_{\text{user}}$, Initial suffix $x_{\text{adv}}^{(0)}$, Harmful target $y_{\text{harm}}$, Safe target $y_{\text{safe}}$, Iterations $T$, Batch size $B$, Top-$K$ candidates, Momentum $\mu$}
\KwOut{Optimized adversarial suffix $x_{\text{adv}}^{*}$ and divergence log $\mathcal{D}$}

Initialize momentum buffer $v \leftarrow 0$ \;
Initialize suffix $x_{\text{adv}} \leftarrow x_{\text{adv}}^{(0)}$ \;
Initialize divergence log $\mathcal{D} \leftarrow \emptyset$ \;

\For{$t = 1$ \KwTo $T$}{
    Construct input $x \leftarrow x_{\text{user}} \oplus x_{\text{adv}}$ \;

    \tcp{Compute dual-precision gradients}
    $g_{\text{P2}} \leftarrow \nabla_{e(x_{\text{adv}})} \mathcal{L} ( f_{\theta}^{\text{P2}}(x), y_{\text{harm}} )$ \;
    $g_{\text{P1}} \leftarrow \nabla_{e(x_{\text{adv}})} \mathcal{L} ( f_{\theta}^{\text{P1}}(x), y_{\text{safe}} )$ \;
    $g_t \leftarrow g_{\text{P1}} + g_{\text{P2}}$ \;

    \tcp{Momentum update}
    $v \leftarrow \mu \cdot v + (1 - \mu) \cdot g_t$ \;

    \tcp{Candidate sampling}
    \For{each token $i$ in $x_{\text{adv}}$}{
        Select Top-$K$ tokens with largest negative gradient $-v^{(i)}$ \;
    }
    Sample batch $\mathcal{X}_{\text{cand}} = \{\tilde{x}_1, \dots, \tilde{x}_B\}$ from Top-$K$ tokens \;

    \tcp{Dual-objective evaluation}
    \For{each candidate $\tilde{x}_b \in \mathcal{X}_{\text{cand}}$}{
        $\mathcal{L}_{\text{total}}^{(b)} \leftarrow \mathcal{L} ( f_{\theta}^{\text{P2}}(x_{\text{user}} \oplus \tilde{x}_b), y_{\text{harm}} ) + \lambda \mathcal{L} ( f_{\theta}^{\text{P1}}(x_{\text{user}} \oplus \tilde{x}_b), y_{\text{safe}} )$ \;
    }

    $x_{\text{adv}} \leftarrow \arg\min_{\tilde{x}_b} \mathcal{L}_{\text{total}}^{(b)}$ \;

    \tcp{Check for precision-induced jailbreak}
    \If{Jailbroken($f_{\theta}^{\text{P2}}(x_{\text{user}} \oplus x_{\text{adv}})$) $\land$ Refused($f_{\theta}^{\text{P1}}(x_{\text{user}} \oplus x_{\text{adv}})$)}{
        $\mathcal{D} \leftarrow \mathcal{D} \cup \{(x_{\text{user}}, x_{\text{adv}}, t)\}$ \tcp*{Record the case and continue iterating}
    }
}
\KwRet{$x_{\text{adv}}, \mathcal{D}$}
\caption{Dual-Precision Momentum-Guided GCG}
\label{alg:dp_gcg}
\end{algorithm}

\subsection{Dual-Precision Differential Testing}
\label{sec:dual}
\label{sec:momentum}

\noindent\textbf{Joint objective formulation.}
We formulate the search for divergence-inducing suffixes as a joint optimization problem across two numerical precision configurations~\cite{zou2023universal,szegedy2013intriguing}.
Let $f_\theta^{(p_1)}$ denote the model executed under precision configuration 1 (e.g., \texttt{bfloat16}) and
$f_\theta^{(p_2)}$ the same model executed under precision configuration 2 (e.g., \texttt{float16}).

Given a user query $x_{\text{user}}$ and a suffix $x_{\text{adv}}$, the full input is
\begin{equation}
x = x_{\text{user}} \oplus x_{\text{adv}}.
\end{equation}

Autoregressive generation can be expressed as minimizing the negative log-likelihood (NLL) of a target sequence.
Let $\mathcal{L}(f(x), y)$ denote the cross-entropy loss between model outputs and a reference sequence $y$.

We optimize the suffix by minimizing a dual-precision objective:

\begin{equation}
\min_{x_{\text{adv}}}
\Big[
\mathcal{L}\big(f_\theta^{(p_2)}(x), y_{\text{harm}}\big)
+
\lambda
\mathcal{L}\big(f_\theta^{(p_1)}(x), y_{\text{safe}}\big)
\Big],
\end{equation}

where $y_{\text{harm}}$ and $y_{\text{safe}}$ are predefined target sequences. In our experiment, we set $\lambda$ to 1 in order to balance the two losses. Jailbreak detection follows the \texttt{test\_prefixes} protocol from GCG~\cite{zou2023universal}. This prefix-based oracle is applied to the first generated sentence, where the affirmative or refusal decision is made, ensuring that cases of ``refusal followed by harmful content'' are correctly identified as safe rather than harmful. By jointly optimizing these terms, we exploit numerical discrepancies between $p_1$ and $p_2$ to find a suffix that remains safe under one precision but triggers jailbreak under the other.

\noindent\textbf{Momentum-guided candidate search.}
Since the suffix $x_{\text{adv}}$ is a discrete token sequence, direct gradient updates are not feasible.
We therefore adopt a coordinate-wise search procedure guided by gradients of the joint objective.

\noindent\textbf{Dual-gradient computation.}
For the $i$-th token in the suffix, we compute gradients of both loss terms with respect to its one-hot embedding representation:
\begin{equation}
g_{p_2}^{(i)} =
\nabla_{\mathbf{x_i}}
\mathcal{L}\big(f_\theta^{(p_2)}(x), y_{\text{harm}}\big),
\end{equation}
\begin{equation}
g_{p_1}^{(i)} =
\nabla_{\mathbf{x_i}}
\mathcal{L}\big(f_\theta^{(p_1)}(x), y_{\text{safe}}\big).
\end{equation}

The two gradients are aggregated to form a joint direction
\begin{equation}
g^{(i)} = g_{p_2}^{(i)} + g_{p_1}^{(i)}.
\end{equation}

\noindent\textbf{Momentum update.}
To stabilize optimization across the non-convex joint landscape, we maintain a momentum buffer~\cite{dong2018boosting,kingma2014adam}
\begin{equation}
v_t = \mu v_{t-1} + (1-\mu) g_t,
\end{equation}
where $\mu$ is the momentum coefficient.

\noindent\textbf{Candidate evaluation.}
Tokens corresponding to the largest negative entries in $v_t$ are selected as Top-$K$ candidates.
We sample a batch of candidate suffixes, compute their exact joint loss
\begin{equation}
\mathcal{L}_{\text{total}} =
\mathcal{L}_{p_2} +
\mathcal{L}_{p_1},
\end{equation}
and retain the suffix with the lowest loss for the next iteration.
The process repeats until the maximum iteration is met.

\subsection{Layer-wise Analysis}
\label{sec:layer}

To investigate where numerical precision differences manifest inside the network, we perform a layer-wise divergence analysis~\cite{meng2022locating,elhage2021mathematical} between two numerically distinct but otherwise identical models (e.g., \texttt{bfloat16} and \texttt{float16}). Both models share the same pretrained weights and are evaluated with an identical input construction and decoding configuration to ensure strict reproducibility. During autoregressive generation, we attach forward hooks to all leaf modules and record the activations from the forward pass that produces the \textit{first} newly generated token. This is the critical decoding step at which the model commits to either an affirmative response (e.g., ``Sure, here is...'') or a refusal, and thus the hidden states at this step are most directly responsible for the divergent output behavior. Analyzing subsequent tokens would conflate two already-diverged generation trajectories, making layer-wise comparison semantically ill-defined.

For each layer $i$, let $O_i^{\text{ref}}$ and $O_i^{\text{dist}}$ denote the output tensors from the reference and distorted models respectively. We first quantify absolute divergence using the Mean Absolute Difference (MAD),
\begin{equation}
\mu_i = \frac{1}{N}\sum_{j=1}^{N}\left| O_{i,j}^{\text{ref}} - O_{i,j}^{\text{dist}} \right|,
\end{equation}
where $N$ is the number of elements in the tensor. While MAD measures the magnitude of discrepancy, it does not indicate where divergence begins to amplify. We therefore define the Relative Divergence Lift (RL) as
\begin{equation}
\mathrm{RL}_i =
\frac{\mu_i - \max_{k<i}\mu_k}
{\max_{k<i}\mu_k + \varepsilon},
\end{equation}
where $\varepsilon$ is a small constant for numerical stability. RL captures the relative increase in divergence compared to all preceding layers, highlighting points where numerical perturbations undergo amplification rather than gradual accumulation.

We identify critical layers using a percentile-based criterion,
\begin{equation}
\mathcal{C} = \{\, i \mid \mathrm{RL}_i > \mathrm{Percentile}_{95}(\mathrm{RL}) \,\},
\end{equation}
which avoids architecture-dependent thresholds and yields a sparse set of amplification points that can serve as actionable targets for selective precision elevation.

\section{Evaluation}

To evaluate PrecisionDiff on the safety alignment verification task, we conduct a series of experiments to answer the following research questions:

\begin{itemize}
\item \textbf{RQ1}: How prevalent is precision-induced behavioral disagreements across different aligned LLMs and precision configurations?

\item \textbf{RQ2}: How effective is PrecisionDiff at exposing precision-induced disagreements compared with existing differential testing techniques?

\item \textbf{RQ3}: How robust is PrecisionDiff under different experimental settings?

\item \textbf{RQ4}: What internal mechanisms contribute to the emergence of precision-induced disagreements?

\end{itemize}

\subsection{Experimental Setup}

We evaluate a diverse set of open-source aligned LLMs, including Llama-2-7b-chat-hf~\cite{touvron2023llama},  Meta-Llama-3-8B~\cite{grattafiori2024llama}, vicuna-7b-v1.5~\cite{chiang2023vicuna},  Mistral-7B-Instruct-v0.2~\cite{Jiang2023Mistral7} and guanaco-7B-HF~\cite{dettmers2023qlora}. We test 50 harmful queries sampled from the AdvBench dataset~\cite{zou2023universal, jia2024improved}, covering diverse categories including violence, illegal activities, and misinformation.

We focus on the most common precision configurations used in deployment, covering standard floating-point formats (\texttt{bfloat16} and \texttt{float16}) and quantized integer formats (\texttt{int16} and \texttt{int8})~\cite{micikevicius2017mixed,qi2025defeating}. For our main experiments, we conduct differential testing between \texttt{bfloat16} and \texttt{float16},  \texttt{bfloat16} and \texttt{int16}, \texttt{int16} and \texttt{int8} to cover different configurations. All other factors (weights, prompts, decoding parameters, seeds) are kept identical to isolate the effect of precision variation. For each harmful query, we optimize a suffix using PrecisionDiff with a maximum of 500 iterations, batch size of 512, and Top-K=256 for candidate selection. We evaluate the resulting prompt under both precision configurations. To isolate precision as the sole variable, we use model.generation and set do\_sample = False to perform greedy sampling~\cite{brown2020language}, eliminating any influence of stochastic decoding. 

A test instance is considered a successful precision-induced jailbreak if the lower-precision model produces a jailbreak while the higher-precision model remains non-jailbroken. For floating-point pairs, we treat \texttt{BF16} as the higher-precision reference, as it is the standard training precision for most open-source LLMs, while \texttt{FP16} is predominantly used for inference. For integer pairs, \texttt{INT16} is naturally the higher-precision format over \texttt{INT8}. This asymmetric definition reflects a realistic deployment risk: models are often evaluated under higher precision during development but deployed using lower precision for efficiency. To assess the effectiveness and efficiency of our detection approach, we define Success Rate as the proportion of test prompts that trigger a precision-induced jailbreak within the maximum iteration budget. Average Iterations refers to the average number of iterations required for the first successful jailbreak among all successful cases, with a lower value indicating higher jailbreak efficiency.

\subsection{RQ1: Prevalence of Precision-Induced Divergence}

\autoref{tab:main_results} summarizes the results. Behavioral divergence emerges consistently across all tested architectures, confirming that precision variation alone can significantly alter model behavior.

\begin{table}[t]
\centering
\caption{Success rates under differential precision testing across multiple precision transitions.}
\label{tab:main_results}
\resizebox{\linewidth}{!}{
\begin{tabular}{lcccccc}
\toprule
\multirow{2}{*}{Model}
& \multicolumn{2}{c}{\texttt{BF16} vs \texttt{FP16}}
& \multicolumn{2}{c}{\texttt{BF16} vs \texttt{INT16}}
& \multicolumn{2}{c}{\texttt{INT16} vs \texttt{INT8}}\\
\cmidrule(lr){2-3} \cmidrule(lr){4-5} \cmidrule(lr){6-7}
& Success & Iter & Success & Iter & Success & Iter\\
\midrule
Guanaco 7B    & 78.0\%  & 83.8  & 84.0\%  & 78.2  & \textbf{100.0\%} & 23.2\\
Llama-2 7B    & 68.0\%  & \textbf{44.7} & 64.0\%  & 24.0  & 98.0\%  & \textbf{9.3}\\
Mistral-7B    & \textbf{96.0\%} & 75.2  & \textbf{94.0\%} & 79.1  & \textbf{100.0\%} & 39.5\\
Vicuna-7B     & 94.0\%  & 48.9  & 91.5\%  & \textbf{14.8} & \textbf{100.0\%} & 10.8\\
\midrule
Average       & \textbf{84.0\%} & 63.2  & \textbf{83.4\%} & 49.0  & \textbf{99.5\%} & \textbf{20.7}\\
\bottomrule
\end{tabular}
}
\end{table}

The results demonstrate high success rates across all tested models and precision pairs, confirming that precision-induced behavioral divergence is pervasive. Even the lowest observed success rate (68.0\% for Llama-2 7B under \texttt{BF16} vs \texttt{FP16}) implies that more than two-thirds of harmful queries can bypass alignment solely by changing inference precision.

Among the three precision transitions, \texttt{INT16} vs \texttt{INT8} is the easiest to exploit (avg.\ 99.5\%, 20.7 iterations), with three out of four models reaching 100.0\% and roughly $3\times$ fewer iterations than the other pairs, indicating that the representational gap directly correlates with divergence severity. \texttt{BF16} vs \texttt{INT16} occupies an intermediate position (83.4\%, 49.0 avg.\ iterations), consistent with its hybrid nature.

Mistral-7B is consistently the most susceptible model (96.0\%, 94.0\%, 100.0\%), likely due to full-precision alignment tightly fitted to specific numerical conditions. Llama-2 7B shows the lowest rates for floating-point pairs yet still reaches 98.0\% for \texttt{INT16} vs \texttt{INT8}, revealing that RLHF-induced robustness is format-specific: a model that resists floating-point shifts may still be highly vulnerable to integer quantization. Vicuna-7B achieves high success rates with low iteration counts, indicating an easily crossed safety boundary, while Guanaco 7B requires more iterations, suggesting a harder boundary. A model that converges rapidly represents a more immediate threat in low-resource attack scenarios.

  \begin{table}[t]
  \centering
  \caption{Per-category success rates and average iterations over representative models and precision settings.}
  \label{tab:category}
  \begin{tabular}{lccc}
  \toprule
  Category & Count & SR & Avg.\ Iter\\
  \midrule
  Cybercrime        & 99  & \textbf{92.9\%} & 41.4\\
  Malware           & 33  & 90.9\%          & 60.6\\
  Fraud             & 99  & 89.9\%          & \textbf{32.6}\\
  Privacy Violation & 33  & 84.8\%          & 51.5\\
  Violence          & 143 & 82.5\%          & 54.8\\
  Hate Speech       & 33  & 75.8\%          & 41.9\\
  Misinformation    & 66  & 71.2\%          & 54.0\\
  Illegal Substances& 44  & 68.2\%          & 62.7\\
  \bottomrule
  \end{tabular}
  \end{table}

To examine whether precision-induced divergence is uniformly distributed across different types of harmful content, we classify the 50 test behaviors into eight categories and report aggregated results in \autoref{tab:category}.
The results reveal a clear hierarchy in precision sensitivity across harm categories. \textit{Cybercrime} (92.9\%), \textit{Malware} (90.9\%), and \textit{Fraud} (89.9\%) are the most susceptible, while \textit{Misinformation} (71.2\%) and \textit{Illegal Substances} (68.2\%) exhibit the lowest success rates, indicating that models maintain stronger alignment for these topics. Notably, \textit{Fraud} converges fastest (32.6 avg. iterations), suggesting that procedural financial content has weaker safety guardrails. One plausible explanation is that technical behaviors (e.g., writing malware, exploiting vulnerabilities, committing cybercrime) are more closely related to legitimate programming and everyday knowledge, resulting in alignment that is more narrowly tuned and thus more susceptible to precision-induced boundary shifts, whereas socially sensitive topics (e.g., hate speech, illegal substances, misinformation) receive broader and more redundant safety coverage during alignment training. The interaction between harm category and precision pair reveals additional structure. Under \texttt{INT16} vs \texttt{INT8}, all categories achieve near-perfect success, confirming that aggressive integer quantization universally degrades alignment. Under the more challenging \texttt{BF16} vs \texttt{FP16}, category differences become pronounced. Cybercrime and Fraud maintain high rates while Misinformation and Illegal Substances drop substantially, demonstrating that alignment robustness to precision variation is content-dependent. At the individual behavior level, the easiest behaviors cluster around technical topics such as ``hacking into a government database'' and ``committing fraud,'' while the hardest behaviors involve ``spreading misinformation,'' ``synthesizing illegal substances,'' and ``acquiring firearms illegally''.

\begin{framed}
\noindent 
\textbf{Answer to RQ1.} Precision-induced behavioral divergence is pervasive across all tested models. By evaluating various precision pairs, we demonstrate that no model is immune to precision-induced jailbreaks. Furthermore, category-level analysis reveals that the severity of divergence correlates with the representational gap, and sensitivity varies significantly by harm type. Specific categories, such as Cybercrime and Malware, show the highest susceptibility to these precision shifts. These results carry heightened implications for embodied AI systems and autonomous LLM agents, where precision-induced jailbreaks can translate into unsafe physical actions or cascading failures in automated pipelines.
\end{framed}

\subsection{RQ2: Effectiveness Compared to Existing Techniques}

\begin{figure}[t]
\centering
\includegraphics[width=\linewidth]{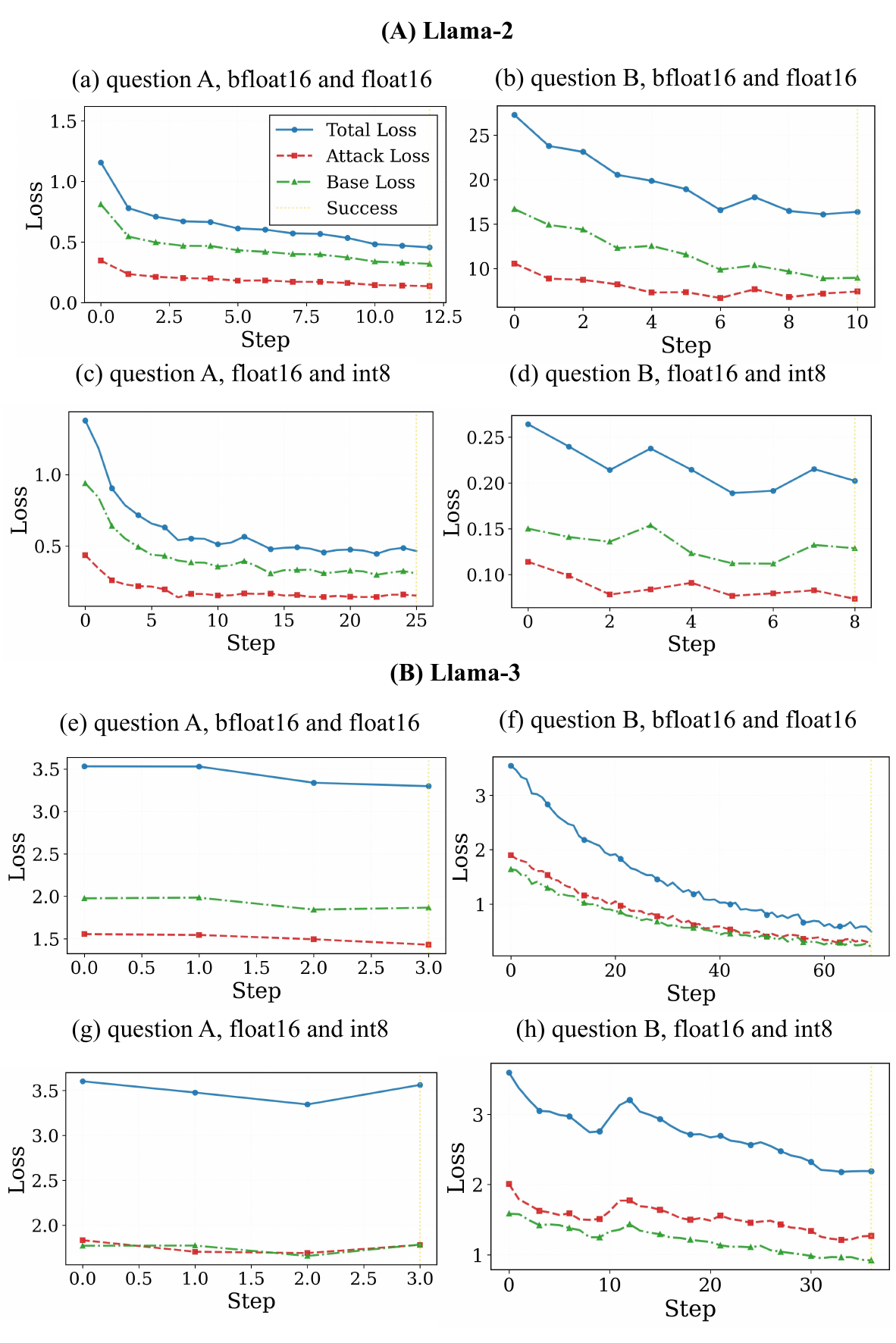}
\caption{
The change of Loss during the Adversarial Prompt Optimizer process.
}
\Description{}
\label{loss}
\end{figure}

\begin{table}[t]
\centering
\caption{Comparison of PrecisionDiff with baseline methods (50 test cases, max\_step=200).}
\label{tab:comparison}
\resizebox{\linewidth}{!}{%
\begin{tabular}{lcccc}
\toprule
& \multicolumn{2}{c}{Vicuna-7B} & \multicolumn{2}{c}{Llama2-7B} \\
\cmidrule(lr){2-3} \cmidrule(lr){4-5}
Method & Success & Iter & Success & Iter \\
\midrule
Random Search            & 2.0\%  & 19.0 & 0.0\%  & N/A \\
Fuzzing (AFL++)          & 8.0\%  & 84.0 & 0.0\%  & N/A \\
Genetic Algorithm        & 8.0\%  & 12.3 & 0.0\%  & N/A \\
Standard GCG (\texttt{FP16}) & 50.0\% & 59.0 & 8.0\%  & 94.3 \\
PrecisionDiff (Ours)     & \textbf{72.0\%} & \textbf{45.0} & \textbf{68.0\%} & \textbf{19.4} \\
\bottomrule
\end{tabular}
}
\end{table}

As shown in \autoref{loss}, the losses decrease smoothly and the testing reaches success without oscillation, indicating a stable optimization trajectory. This smooth convergence contrasts with the instability typically observed in single-precision jailbreak optimization, where the suffix can overshoot the decision boundary and require costly recovery steps; the dual-objective formulation in PrecisionDiff effectively acts as a regularizer that keeps the search near the precision-sensitive boundary rather than crossing into strongly jailbroken territory. To evaluate the effectiveness of PrecisionDiff, we compare PrecisionDiff against standard GCG, random search, fuzzing, and genetic algorithm baselines on Vicuna 7B using 50 test cases (\autoref{tab:comparison}).

The results reveal a clear hierarchy among the methods. Conventional differential testing approaches are largely ineffective at discovering precision-sensitive divergences. Fuzzing (AFL++) and genetic algorithm both achieve only 8.0\%, while random search reaches 2.0\% with an average of 19.0 iterations. These low success rates confirm that precision-sensitive divergences occupy a vanishingly narrow region in the high-dimensional input space, which undirected exploration strategies cannot reliably reach. The discrete nature of LLM token spaces further compounds this difficulty, as mutation operators designed for continuous domains lose their effectiveness when applied to token sequences: single-token substitutions or crossover operations are unlikely to land in the precision-sensitive boundary zone by chance. Notably, genetic algorithm achieves the same 8.0\% as fuzzing despite its structured crossover and selection operators, because the fitness landscape of precision-induced divergence is essentially flat outside the narrow boundary zone, causing the population to drift without meaningful selective pressure and converge to non-divergent inputs.

Standard GCG achieves 50.0\% (59.0 avg.\ iterations) but can only occasionally encounter precision-sensitive boundary regions, and once the optimization crosses such a boundary, subsequent iterations tend to become ineffective. In contrast, PrecisionDiff explicitly targets these boundary regions via dual-objective optimization, allowing it to systematically expose a greater number of divergence points. The key insight is that standard GCG optimizes for jailbreak success under a single precision, which leads the search toward the center of the jailbreak region rather than its precision-sensitive fringe. PrecisionDiff's simultaneous safe-objective on the reference precision acts as a repulsive force that continuously pushes the suffix toward the narrow transition zone between safe and unsafe behavior.

PrecisionDiff achieves 72.0\% success with only 45.0 avg.\ iterations on Vicuna-7B, representing a $1.4\times$ success rate gain and $1.3\times$ fewer iterations than standard GCG. The advantage is even more pronounced on Llama-2-7B, where PrecisionDiff reaches 68.0\% against standard GCG's 8.0\% ($8.5\times$ improvement), highlighting that dual-precision gradient guidance is especially indispensable for strongly aligned models where single-precision optimization fails to locate the narrow precision-sensitive boundary.

\noindent\textbf{Time overhead analysis.}
We further measure the per-iteration wall-clock time of PrecisionDiff and standard GCG on NVIDIA A40 GPUs (Guanaco-7B, batch size = 256, Top-K = 256).
Standard GCG, which loads one model in \texttt{FP16} and performs a single round of gradient computation, candidate sampling, loss evaluation, and generation checking, takes     5.31\,s per iteration on a single A40.
PrecisionDiff loads two model copies (\texttt{BF16} and \texttt{FP16}) on two A40 GPUs and performs these operations on both models sequentially, resulting in 9.97\,s per iteration ($1.88\times$ overhead).
Despite the roughly doubled per-step cost, PrecisionDiff requires $1.3\times$ fewer iterations to succeed (45.0 vs.\ 59.0), yielding a comparable total wall-clock time ($45.0 \times 9.97 \approx 449$\,s vs.\ $59.0 \times 5.31 \approx 313$\,s) while achieving a substantially higher success rate (72.0\% vs.\ 50.0\%). From a practical standpoint, the additional GPU required for the second model copy represents a modest and well-justified overhead. In a typical safety evaluation pipeline, the improved detection coverage is far more valuable than the marginal compute savings of single-precision testing, particularly given that undetected precision-induced vulnerabilities may persist silently in deployed systems.

\begin{framed}
\noindent 
\textbf{Answer to RQ2.} PrecisionDiff detects precision-induced divergence efficiently and outperforms all baselines. By specifically targeting precision shifts rather than generic jailbreaks, PrecisionDiff achieves significantly higher success rates with fewer iterations than conventional random search or standard GCG. Furthermore, while the per-iteration time overhead is higher, it is well compensated by the reduced total iteration count. Ultimately, PrecisionDiff provides the best overall detection performance.
\end{framed}

\subsection{RQ3: Effectiveness in Different Settings}

To assess the generalizability of PrecisionDiff, we evaluate its effectiveness under two alternative experimental settings: (1) stochastic decoding with do\_sample=True, and (2) a reduced optimization budget with max\_step=200. These settings are practically relevant: stochastic decoding is the default mode in many chatbot deployments, while budget-constrained testing reflects scenarios where rapid safety evaluation is required prior to deployment or where compute resources are shared across multiple models.

\begin{table}[t]
\centering
\caption{Success rates under differential precision testing across multiple precision transitions when do\_sample=True.}
\label{tab:sample_true}
\begin{tabular}{lcccc}
\toprule
& \multicolumn{2}{c}{\textbf{Llama-3 8B}} & \multicolumn{2}{c}{\textbf{Llama-2 7B}} \\
\cmidrule(lr){2-3} \cmidrule(lr){4-5}
\textbf{Precision Transition} & Success & Avg. Iter & Success & Avg. Iter \\
\midrule
BF16 vs FP16 & \textbf{100.0\%} & 87.7 & 90.0\% & \textbf{11.3} \\
BF16 vs INT16 & \textbf{98.0\%} & 109.2 & 95.8\% & \textbf{15.4} \\
INT16 vs INT8 & \textbf{96.0\%} & 71.5 & 90.0\% & \textbf{13.8} \\
\bottomrule
\end{tabular}
\end{table}

\begin{table}[t]
\centering
\caption{Success rates under differential precision testing across multiple precision transitions when max\_step=200.}
\label{tab:max200}
\begin{tabular}{lcccc}
\toprule
& \multicolumn{2}{c}{\textbf{BF16 vs FP16}} & \multicolumn{2}{c}{\textbf{FP16 vs BF16}} \\
\cmidrule(lr){2-3} \cmidrule(lr){4-5}
Model & Success & Avg. Iter & Success & Avg. Iter \\
\midrule
Guanaco 7B & 62.0\% & 73.0 & 52.0\% & 81.2 \\
Llama-2 7B & 68.0\% & 19.4 & 72.0\% & 24.2 \\
Mistral-7B & \textbf{92.0\%} & 55.7 & 72.0\% & 76.2 \\
Vicuna-7B & 72.0\% & 45.0 & 62.0\% & 48.7 \\
\bottomrule
\end{tabular}
\end{table}

We first examine the effect of stochastic decoding. Our main experiments use greedy decoding to isolate precision as the sole variable. Here we enable do\_sample=True and test Llama-3 8B and Llama-2 7B across three precision transitions (\autoref{tab:sample_true}).

Precision-induced divergence remains prevalent under stochastic decoding. Llama-3 8B achieves 100.0\%/98.0\%/96.0\% and Llama-2 7B achieves 90.0\%/95.8\%/90.0\% across the three transitions, generally higher than greedy decoding, since sampling randomness amplifies small logit differences caused by precision variation. Notably, the two models show contrasting efficiency. Llama-2 7B converges rapidly (avg.\ 11--15 iterations) while Llama-3 8B requires considerably more effort (71--109 iterations), indicating that Llama-3's alignment provides greater resistance to adversarial optimization despite ultimately yielding higher success rates. This trade-off between convergence speed and final success rate suggests that stronger alignment delays but does not prevent precision-induced divergence. Sufficiently optimized suffixes can still locate the precision-sensitive boundary, but the path to that boundary is longer. The interaction between stochastic decoding and precision variation thus compounds alignment fragility in a previously uncharacterized way, as both sources of stochasticity (numerical rounding and sampling) conspire to widen the effective jailbreak region.

We next examine PrecisionDiff under a reduced optimization budget. Reducing max\_step from 500 to 200 (a 60\% reduction) tests the practical efficiency of our approach (\autoref{tab:max200}). This setting also allows us to compare bidirectional transitions (\texttt{BF16} not jailbreak vs \texttt{FP16} jailbreak and \texttt{FP16} not jailbreak vs \texttt{BF16} jailbreak) across four model configurations.

With max\_step=200, success rates remain substantial (62.0\%--92.0\% for \texttt{BF16} vs \texttt{FP16} and 52.0\%--90.0\% for \texttt{FP16} vs \texttt{BF16}). The \texttt{BF16}$\to$\texttt{FP16} direction consistently yields higher success rates than the reverse (e.g., Mistral-7B at 92.0\% vs 72.0\%), and iteration counts are generally higher in the reverse direction (e.g., Mistral-7B goes from 55.7 to 76.2). This asymmetry is consistent with the decision boundary shift observed in RQ4: because most models are aligned under \texttt{BF16} training precision, the safety boundary is calibrated to \texttt{BF16}'s numerical characteristics. Switching the reference model to \texttt{FP16} therefore requires more optimization effort to cross a boundary that was not shaped by its precision. In other words, \texttt{BF16} naturally sits closer to the safe side of its own aligned boundary, making it easier to keep safe under \texttt{BF16} while inducing jailbreaks under \texttt{FP16}. Across the board, success rates remain above 50\% even with a 60\% reduction in the optimization budget, underscoring the practical efficiency of PrecisionDiff.

\begin{framed}
\noindent 
\textbf{Answer to RQ3.} PrecisionDiff generalizes effectively across diverse configurations. Under varying conditions, such as stochastic decoding and significantly reduced optimization budgets, the framework consistently maintains robust detection success rates. Furthermore, bidirectional testing reveals an asymmetric pattern in precision sensitivity. Transitions between specific formats, such as \texttt{BF16} and \texttt{FP16}, generally yield the most pronounced divergence behaviors.
\end{framed}

\subsection{RQ4: Contributing Factors and Internal Mechanisms}

To understand what drives precision-induced behavioral divergence, we investigate two complementary aspects: how precision changes affect the macroscopic decision boundary, and where divergence amplification occurs across network layers.

\begin{figure}[!t]
    \centering
    \includegraphics[width=\linewidth]{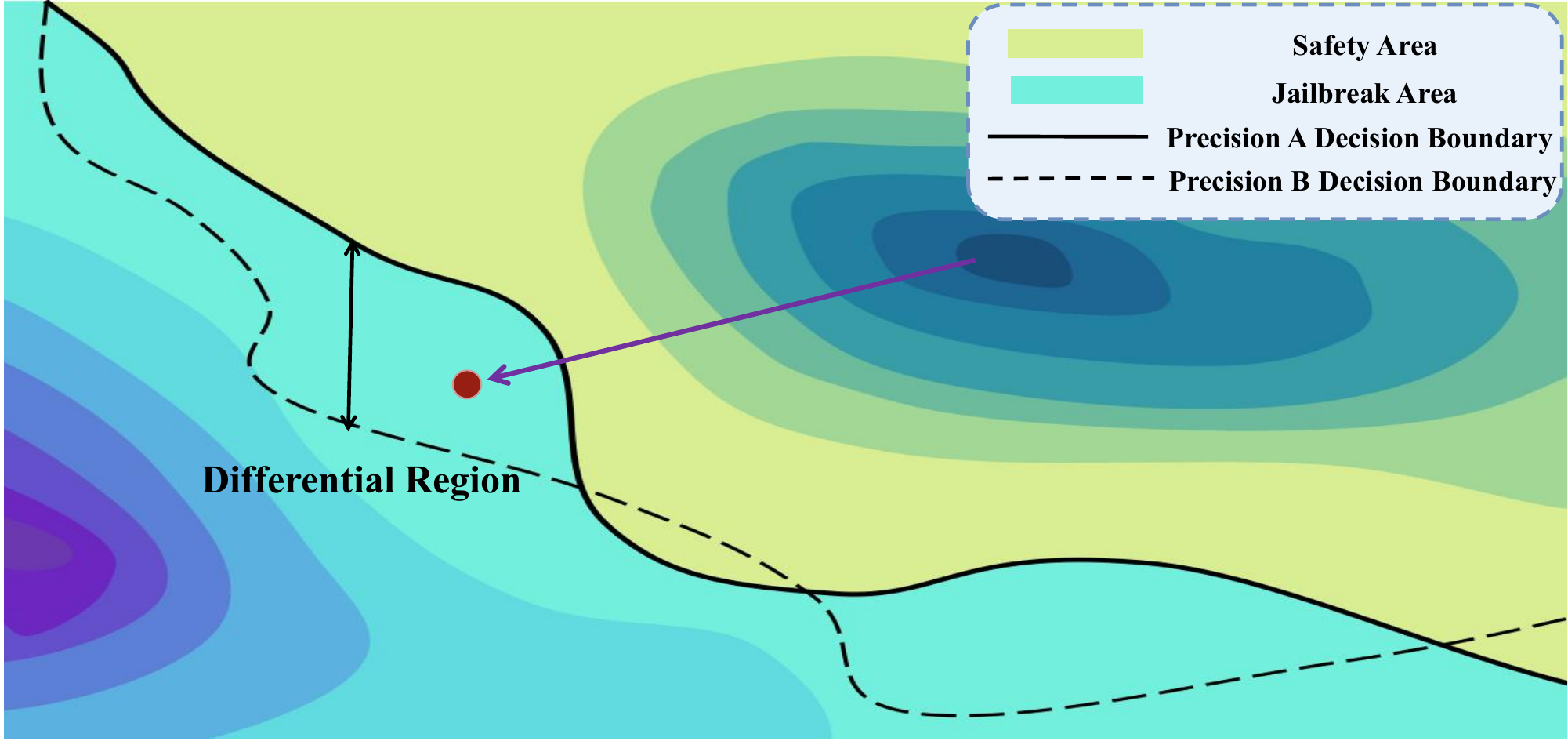}
    \caption{
        Jailbreak decision boundaries of large language models under different precision configurations.
    }
    \Description{}
    \label{decision}
\end{figure}

\noindent\textbf{Decision boundary shift.}
As illustrated in \autoref{decision}, precision changes systematically deform the safety boundary rather than introducing random noise, effectively enlarging the jailbreakable region in representation space. This geometric shift explains the asymmetric pattern in \autoref{tab:max200}. \texttt{BF16}$\to$\texttt{FP16} generally yields higher rates than the reverse (e.g., Mistral-7B at 92.0\% vs 72.0\%), since the boundary shifts closer to the jailbreak side under one precision. The alignment procedure also influences robustness. Guanaco, trained with QLoRA-based 4-bit quantization, consistently shows the lowest success rates (78.0\% in \autoref{tab:main_results}), suggesting that quantization-aware alignment may provide partial cross-precision robustness. In contrast, Mistral-7B with standard full-precision alignment exhibits the highest sensitivity. Models aligned under quantized conditions may inadvertently develop more robust safety boundaries, suggesting that quantization-aware alignment could be a viable strategy for improving cross-precision reliability. As a control, loading both copies under the \emph{same} precision yields bit-for-bit identical outputs, confirming that divergences are caused exclusively by precision differences.

\begin{figure}[t]
    \centering
    \includegraphics[width=\linewidth]{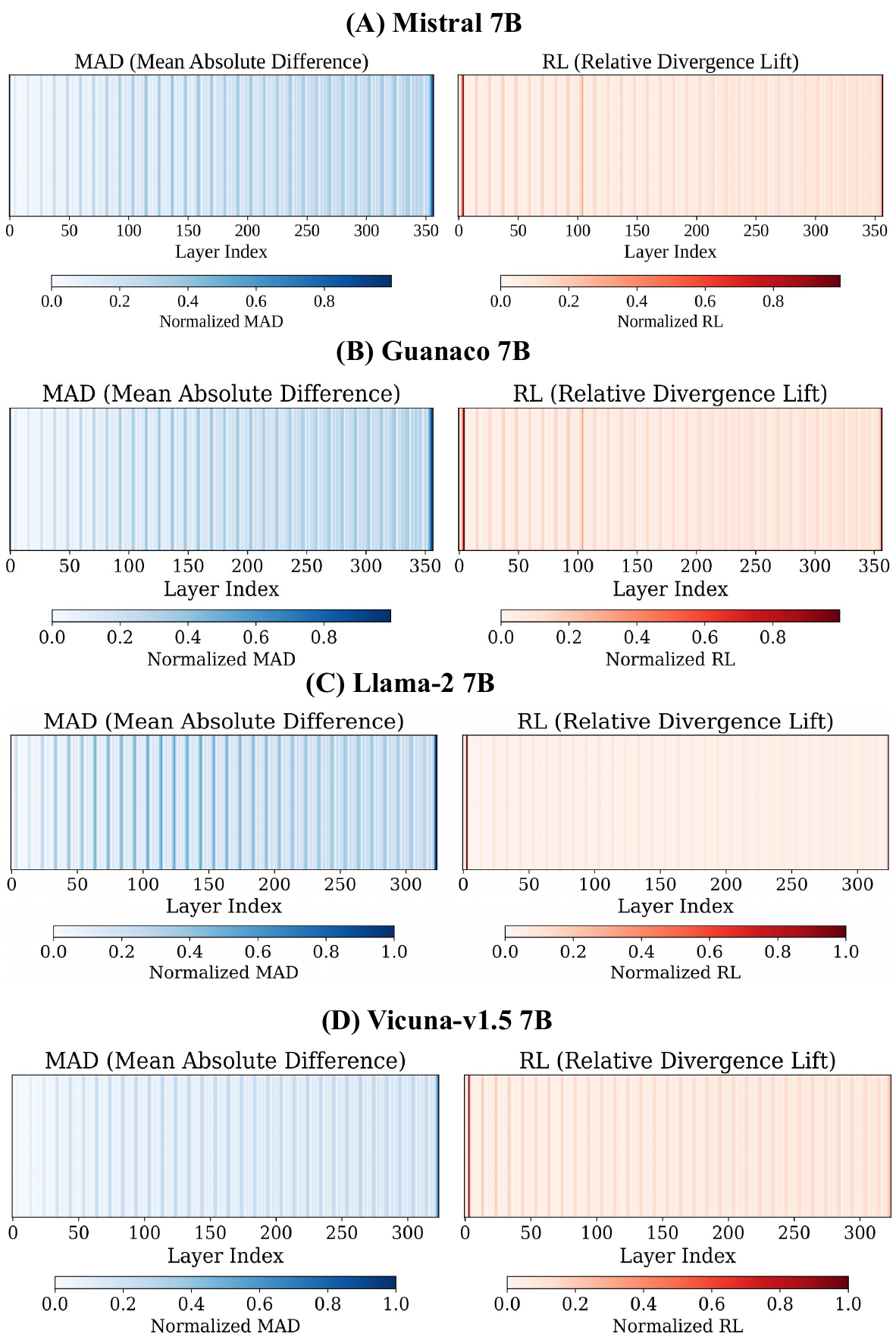}
    \caption{
The layer-wise source localization of disagreements on four open-source models between different precisions.
    }
    \Description{}
    \label{layer}
\end{figure}

\noindent\textbf{Layer-wise divergence localization.}
To pinpoint where divergence amplifies inside the network, we apply the layer-wise analysis methodology described in Section~3.4 to the divergence-inducing inputs discovered by PrecisionDiff in Section~4.2. As shown in \autoref{layer}, the divergence does not grow monotonically with depth but follows a structured pattern. Although small discrepancies are present across most layers, only a small subset exhibits pronounced Relative Divergence Lift (RL) peaks, and these peaks consistently appear at the input-stage layers ($\text{Layer}_0$), the output-end modules ($\text{Layer}_{\text{norm}}$ and $\text{Layer}_{\text{head}}$), and the attention mechanisms ($\mathbf{W}_Q, \mathbf{W}_K$) within the initial blocks.

This behavior can be explained by the numerical properties of these operations. The input-stage layer ($\text{Layer}_0$) directly conditions all downstream representations, so even minor rounding errors introduced at this stage propagate and amplify throughout the network. Attention projection operators ($\mathbf{W}_Q, \mathbf{W}_K$) involve large matrix multiplications where rounding differences accumulate. The resulting deviations in attention logits are further amplified by the subsequent softmax computation, which is sensitive to small input perturbations. Because softmax is an exponential function, even small additive differences in attention logit space are mapped to multiplicative differences in attention weight space. This causes the model to assign substantially different attention patterns across token positions under the two precisions, propagating the divergence into all subsequent residual stream computations. Output-end modules, namely the normalization layer ($\text{Layer}_{\text{norm}}$) and the head layer ($\text{Layer}_{\text{head}}$), are also numerically sensitive. Normalization relies on estimating per-token statistics whose variance computation is prone to rounding errors under reduced-precision arithmetic. The head layer maps the final hidden state directly to logits, so any accumulated deviation is directly reflected in the output distribution.

These findings indicate that precision-induced divergence is not uniformly distributed but concentrated at the input-stage layers, output-end modules, and attention mechanisms within the initial blocks, providing actionable targets for selective precision elevation. Concretely, a mitigation strategy that maintains higher precision (e.g., \texttt{FP32} or \texttt{BF16}) exclusively for $\text{Layer}_0$, $\mathbf{W}_Q$, $\mathbf{W}_K$, $\text{Layer}_{\text{norm}}$, and $\text{Layer}_{\text{head}}$ while aggressively quantizing intermediate layers could in principle suppress the majority of divergence amplification with minimal impact on overall inference throughput. The consistency of this pattern across all four tested architectures further suggests that these amplification hotspots are a structural property of the transformer design rather than a model-specific artifact.

\begin{framed}
\noindent 
\textbf{Answer to RQ4.} Precision-induced divergence is driven by systematic deformations rather than random noise. Precision changes systematically alter the safety decision boundary, producing clear asymmetric divergence patterns. Furthermore, this divergence is not uniformly distributed across the network. The amplification is primarily concentrated at the input-stage layers, specific attention mechanisms, and output-end modules, which complement each other in driving the alignment failure.
\end{framed}

\section{Threats to Validity}

Our experimental design controls for numerical precision as the sole variable by maintaining identical model weights, inputs, and decoding configurations across precision variants. First, the choice of optimization hyperparameters (momentum coefficient, batch size, Top-K selection) in PrecisionDiff may influence the success rate and convergence speed. We mitigated this by conducting preliminary tuning experiments and selecting values that demonstrated stable performance across multiple models.

Our evaluation focuses on open-source aligned LLMs (Llama-2, Llama-3, Mistral, Vicuna, Guanaco) with publicly available weights. The generalizability of our findings to proprietary models (e.g., GPT-4, Claude) or models with different alignment procedures remains to be validated. Additionally, we primarily investigate the \texttt{bfloat16}-\texttt{float16} precision pair. While we also consider \texttt{int16} and \texttt{int8} quantization, the full spectrum of mixed-precision configurations (e.g., per-layer mixed precision, dynamic quantization) requires further investigation.

We rely on automated classifiers to determine whether a model output constitutes a jailbreak or refusal. While these classifiers have been validated in prior work, they may introduce false positives or false negatives, particularly for borderline cases. Additionally, our definition of precision-induced jailbreak focuses on asymmetric divergence (safe under one precision, unsafe under another), which may not capture all forms of behavioral disagreements relevant to deployment safety.

\section{Related Work}

\subsection{LLM Jailbreak Attacks}

A growing body of work studies jailbreak behaviors in aligned LLMs.
Prompt-based attacks~\cite{wei2023jailbroken,shen2024anything} craft natural language instructions that bypass refusal heuristics, while gradient-based methods such as Greedy Coordinate Gradient (GCG)~\cite{zou2023universal,jia2024improved} and automated prompt optimization frameworks~\cite{liu2023autodan,jones2023automatically,chao2025jailbreaking,carlini2023aligned} explicitly optimize token sequences to maximize the likelihood of target responses.

These approaches treat the model as a fixed function and search for adversarial prompts within the input space.
In contrast, our work identifies a previously unexplored precision-sensitive dimension, namely numerical precision configurations.
We demonstrate that behavioral divergence can arise even with a fixed input, driven solely by floating-point precision variations during inference, without relying on semantic manipulation.

Existing jailbreak methods operate under a single precision setting and thus discover inputs with consistent cross-precision behaviors. PrecisionDiff instead targets decision boundary gaps between precision variants, exposing inputs that are safe under one precision but exploitable under another. To our knowledge, it is the first automated differential testing framework for precision-induced jailbreaks in aligned LLMs. Fine-tuning can also compromise alignment~\cite{qi2023fine,bianchi2023safety}, further highlighting the fragility of safety mechanisms. The threat surface of jailbreak attacks extends beyond text-based assistants. Embodied AI systems~\cite{zhang2024badrobot} and autonomous LLM agents~\cite{zhang2025breaking} are particularly susceptible, as jailbreak-induced failures can trigger unsafe physical actions or amplify malfunction cascades in automated pipelines.

\subsection{System-Level Effects on Model Behavior}

Prior research has established that implementation-level details (such as compiler optimizations~\cite{chen2025your}, hardware architectures~\cite{moller2026hardware}, algebra backends~\cite{moller2025adversarial}, and deployment pipelines~\cite{guo2019empirical}) can introduce measurable discrepancies in model outputs. However, existing studies remain largely confined to smaller vision models or numerical reproducibility. This work bridges software engineering and AI safety by demonstrating that precision variations cause qualitative shifts in safety behavior, not merely performance trade-offs. These findings highlight the need to treat system-level design choices as critical factors in ensuring reliable and consistent model alignment.

\subsection{Differential Testing in SE4AI}

Differential testing has emerged as an effective methodology for detecting inconsistencies and precision-sensitive behaviors in AI systems by systematically comparing the behaviors of multiple model variants under identical inputs~\cite{pham2019cradle,xie2019diffchaser}.
Its core principle is to expose discrepancies arising from implementation differences, configuration variations, or environmental factors, to reveal hidden bugs, robustness issues, or security issues~\cite{pham2019cradle,xie2019diffchaser,guo2020audee}.

Prior work has applied differential testing to discover inconsistencies across different model architectures~\cite{xie2019diffchaser,zhan2025allclose}, training configurations~\cite{wang2024d}, software frameworks~\cite{pham2019cradle,guo2020audee}, and hardware backends~\cite{yan2025graphfuzz}.
These approaches have proven effective for traditional supervised learning tasks, where divergence is measured via discrete prediction disagreements.
Recent advances have integrated neural architecture fuzzing~\cite{gu2022muffin,yang2025may} and large language models~\cite{deng2023large,li2024enhancing} to enhance test generation and bug detection in deep learning frameworks~\cite{zhang2025your,yu2025towards}, further improving coverage and effectiveness.

However, conventional differential testing tools such as fuzzing~\cite{deng2023large,gu2022muffin} and genetic algorithms~\cite{yan2025graphfuzz} exhibit extremely low success rates when applied to detect precision-induced inconsistencies in aligned LLMs. The fundamental challenge is that precision-sensitive divergences occupy a vanishingly small region of the high-dimensional input space, making them difficult to locate through unguided search.

Our work extends differential testing to a new dimension, namely numerical precision configurations in aligned large language models.
Unlike prior approaches that compare different models or implementations~\cite{xie2019diffchaser,zhan2025allclose,wang2024d}, PrecisionDiff performs differential testing across precision variants of the same model during inference, isolating numerical precision as the sole causal factor.
While differential testing is an established paradigm~\cite{pham2019cradle,xie2019diffchaser}, its application to precision-induced jailbreaks in aligned LLMs introduces two non-trivial challenges that demand novel technical solutions.
First, prior differential testing targets classification inconsistencies on continuous or structured inputs~\cite{pham2019cradle,guo2020audee,yan2025graphfuzz}, whereas this work targets \emph{safety-semantic divergence} in the discrete token space of LLMs.
Second, existing approaches operate over continuous input domains~\cite{pham2019cradle,guo2020audee} or structured mutation spaces~\cite{gu2022muffin,yang2025may}, while PrecisionDiff must optimize over a combinatorially large discrete token space where gradient signals are inherently non-differentiable, necessitating the dual-precision GCG formulation.

\section{Conclusion}

We presented PrecisionDiff, the first automated differential testing framework for detecting precision-induced behavioral inconsistencies in LLMs. The framework is general and can be applied to different behavioral objectives and precision configurations. As a case study, we used PrecisionDiff for safety alignment verification, revealing jailbreak divergence caused by precision differences. Experiments on five aligned LLMs show that such inconsistencies are common, with detection rates up to 100\% in some precision transitions. Layer-wise analysis indicates that divergence is concentrated in a small set of precision-sensitive layers, providing actionable guidance for precision-robust deployment. Importantly, PrecisionDiff can automatically generate precision-sensitive test inputs, enabling effective pre-deployment evaluation and supporting improved precision robustness during training. These results highlight numerical precision as an overlooked source of behavioral disagreement in LLMs, with important implications for embodied AI systems and autonomous agents.

\paragraph{Statement of Ethics}
PrecisionDiff is designed to promote responsible LLM deployment by identifying behavioral inconsistencies arising from numerical precision variation. It enables developers to audit model behavior across heterogeneous precision configurations prior to deployment, without using or generating any private user data.

\section{Data Availability}
To facilitate reproducibility while preserving the double-blind review process, the source code of PrecisionDiff, experimental results, and the evaluation datasets (including 50 harmful queries from AdvBench) are hosted in an anonymized repository at \url{https://doi.org/10.5281/zenodo.19250143}. This repository provides an preview of the implementation and results. The non-anonymized source code will be made publicly available on GitHub upon the paper’s official acceptance.

\balance
\bibliographystyle{ACM-Reference-Format}
\bibliography{software}

\end{document}